\documentclass{article}
\usepackage{spconf,amsmath,epsfig,graphicx,subfigure}

\let\OLDthebibliography\thebibliography
\renewcommand\thebibliography[1]{
  \OLDthebibliography{#1}
  \setlength{\parskip}{0pt}
  \setlength{\itemsep}{0pt plus 0.3ex}
}

\begin{document}\sloppy

% Example definitions.
% --------------------
\def\x{{\mathbf x}}
\def\L{{\cal L}}

% Title.
% ------
\title{RailYolact - A Yolact Focused on edge for Real-Time Rail Segmentation}
%
% Single address.
% ---------------
\name{Qihao Qian}
%Address and e-mail should NOT be added in the submission paper. They should be present only in the camera ready paper. 
\address{}

\maketitle

\begin{abstract}
Ensuring obstacle avoidance on the rail surface is crucial for the safety of autonomous driving trains and its first step is to segment the regions of the rail. We chose to build upon Yolact for our work. To address the issue of rough edge in the rail masks predicted by the model, we incorporated the edge information extracted by edge operator into the original Yolact's loss function to emphasize the model's focus on rail edges. Additionally, we applied box filter to smooth the jagged ground truth mask edges cause by linear interpolation. Since the integration of edge information and smooth process only occurred during the training process, the inference speed of the model remained unaffected. The experiments results on our custom rail dataset demonstrated an improvement in the prediction accuracy. Moreover, the results on Cityscapes showed a 4.1 and 4.6 improvement in $AP$ and $AP_{50}$ , respectively, compared to Yolact.
\end{abstract}
\section{Introduction}
Automatic train driving is a significant research area in modern intelligent rail transportation systems. Obstacle avoidance on the rail surface is particularly important for ensuring the safe operation of trains. To address this task, a common approach is to utilize laser radar and high-definition, high-frame-rate cameras as infrastructure for data acquisition and fusion, enabling the detection of obstacles in the train's operating area and timely action. This method can be further divided into several major steps: rail segmentation, fusion of image and radar data, and obstacle detection. This paper primarily focuses on the first step in this approach—rail segmentation. We work on how to utilize deep learning based instance segmentation algorithms to accurately and quickly segment the rail regions in the collected videos.

Considering the high speed of trains, it is crucial to quickly verify the presence of obstacles in the forward operating area to make timely decisions and ensure train safety. And if the algorithm incorrectly segment the rails, it will result in inaccurate warnings and erroneous decision-making. As a result, it is especially important to achieve real-time and accurate segmentation of the rail region. To achieve real-time performance, the segmentation algorithm should operate at a speed of no less than 20 frames per second (fps). Existing instance segmentation algorithms can be broadly categorized into two-stage algorithms, represented by Mask R-CNN~\cite{he2017mask}, and one-stage algorithms based on modified one-stage object detectors like RetinaNet~\cite{lin2017focal}. Two-stage instance segmentation algorithms provide high segmentation accuracy but typically have inference speeds of less than 10 fps, which cannot meet the real-time requirements of practical applications. On the other hand, although single-stage instance segmentation algorithms meet the speed requirements, the segmentation masks they generate are often coarse, particularly when it comes to predicting rail edges. Achieving an inference speed of over 20 fps with two-stage algorithms poses significant challenges. However, for relatively simple tasks like rail segmentation, it is feasible to enhance single-stage algorithms to meet the accuracy requirements.

\begin{figure}[ht]
\begin{minipage}[b]{1.0\linewidth}
  \centering
\centerline{\includegraphics[width=8.5cm]{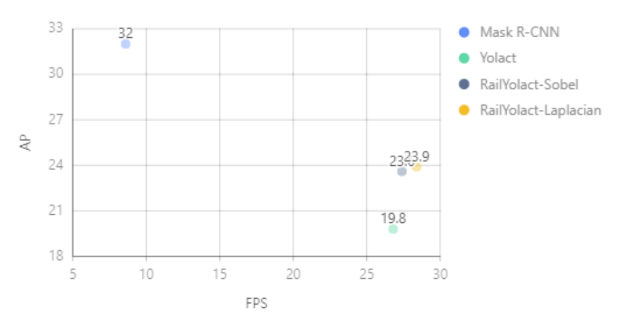}}
  % \centerline{(a) Result 1}\medskip
\end{minipage}
\caption{Comparison on FPS and AP on Cityscapes~\cite{cordts2016cityscapes}}
\end{figure}

Due to the train always being directly above the rail and the camera is installed at the front of the train, rail captured by the camera always show as a trapezoid. By constructing a network pathway that reinforces the geometric structure learning of the aforementioned regular description, a single-stage instance segmentation algorithm can focus more on the rail's edge information during the training process, enhancing its ability to segment rail with greater accuracy. After conducting multiple experiments, we selected the Yolact~\cite{bolya2019yolact} model as a basis for improvement due to its superior performance, aiming to enhance its focus on instance edge information for our rail segmentation task. Our main approach involves adding an edge detection branch at the final step of the model's mask generation. We use traditional edge detection operators to extract the edge information of the rail from both the predicted and ground truth masks, and then calculate the loss of the edge information that is incorporated into the loss function. This enables the model to acquire a more comprehensive understanding of the rail's edge information. The edge detection operator can be implemented as a convolutional kernel without learnable parameters, making our method simple and efficient. Additionally, directly generating high-resolution masks significantly increases the network's parameter size, leading to a decrease in algorithm speed. Existing approaches commonly employ linear interpolation to downscale the ground truth masks, aligning them with the predicted masks generated by the algorithm. However, these approaches introduces jagged edges to the ground truth masks, which learned by the network, resulting in the prediction of jagged masks. We propose a smoothing strategy based on box filters to address the issue of jagged mask edges. Since edge information processing only occurs during model training, our method does not affect the model's inference speed.

\section{related work}
\subsection{Instance Segmentation}
Instance segmentation is a task in the field of computer vision that aims to assign each pixel in an image to a specific object instance. Unlike semantic segmentation, which focuses on identifying different object categories in an image, instance segmentation goes a step further by assigning a unique label to each instance. With the emergence of deep learning, particularly the successful application of convolutional neural networks~\cite{lecun1989backpropagation}, significant advancements have been made in instance segmentation. In 2014, Girshick et al. introduced the classic object detection method, Faster R-CNN~\cite{ren2015faster}, which laid the groundwork for instance segmentation. In 2017, He et al. further extended Faster R-CNN and proposed Mask R-CNN~\cite{he2017mask}, which combines object detection with semantic segmentation to achieve accurate pixel-level instance segmentation.
\subsection{Yolact}
To improve the speed of instance segmentation tasks, Daniel Bolya et al. proposed the Yolact~\cite{bolya2019yolact}, which extends the single-stage detector RetinaNet~\cite{lin2017focal} by adding a third branch to predict the combination coefficients of prototypes. Additionally, it generates a set of prototypes from the image features extracted by the backbone using FCN~\cite{long2015fully}. Yolact also introduces a Fast NMS algorithm to replace traditional NMS, which accelerates the overall inference speed with minimal loss in accuracy. Finally, for each instance surviving NMS, the coefficients of prototypes are combined with the prototypes through a simple matrix multiplication and activated by a non-linear function Sigmoid, resulting in the generation of masks for each instance in the image.
\subsection{Edge Detection}
Image edge processing has been a prominent research area in computer vision for an extensive period. Edge detection encompasses multiple methods, with the most common one leveraging the distinct local feature variations and abrupt gray scale changes in image pixels. In 1968, Irwin Sobel proposed the Sobel operator in his paper "An Isotropic 3x3 Image Gradient Operator" to detect edges. However, the Sobel filter can only detect edges in the horizontal or vertical directions, which was addressed by the Laplacian operator. In the paper  “Faster training of Mask R-CNN by Focusing on Instance Boundaries~\cite{zimmermann2019faster}", the authors introduced the concept of edge information into Mask-R CNN for the first time, resulting in a substantial improvement in the model's performance.

\section{RailYolact}
\subsection{Edge Operator}
\subsubsection{Sobel Operator}
The Sobel operator is a two-dimensional filter used for edge detection, and it can be implemented with a fixed convolution kernel as shown in Equation (1). By applying the $S_x$ and $S_y$ operators to each pixel in the mask, the horizontal and vertical gradient values within the mask can be calculated, respectively. Combining these gradient values from both directions allows us to determine the edge strength and direction of each pixel.
\begin{eqnarray}
S_x=\begin{bmatrix}1&0&-1\\2&0&-2\\1&0&-1\end{bmatrix},S_y=\begin{bmatrix}1&2&1\\0&0&0\\-1&-2&-1\end{bmatrix}
\end{eqnarray}
Figure 2 presents the outcomes of edge detection using the $S_x$ and $S_y$ operators. 

\begin{figure}[ht]
  \begin{minipage}[b]{\linewidth}
    \centering
    \subfigure[car]{\includegraphics[width=0.3\linewidth]{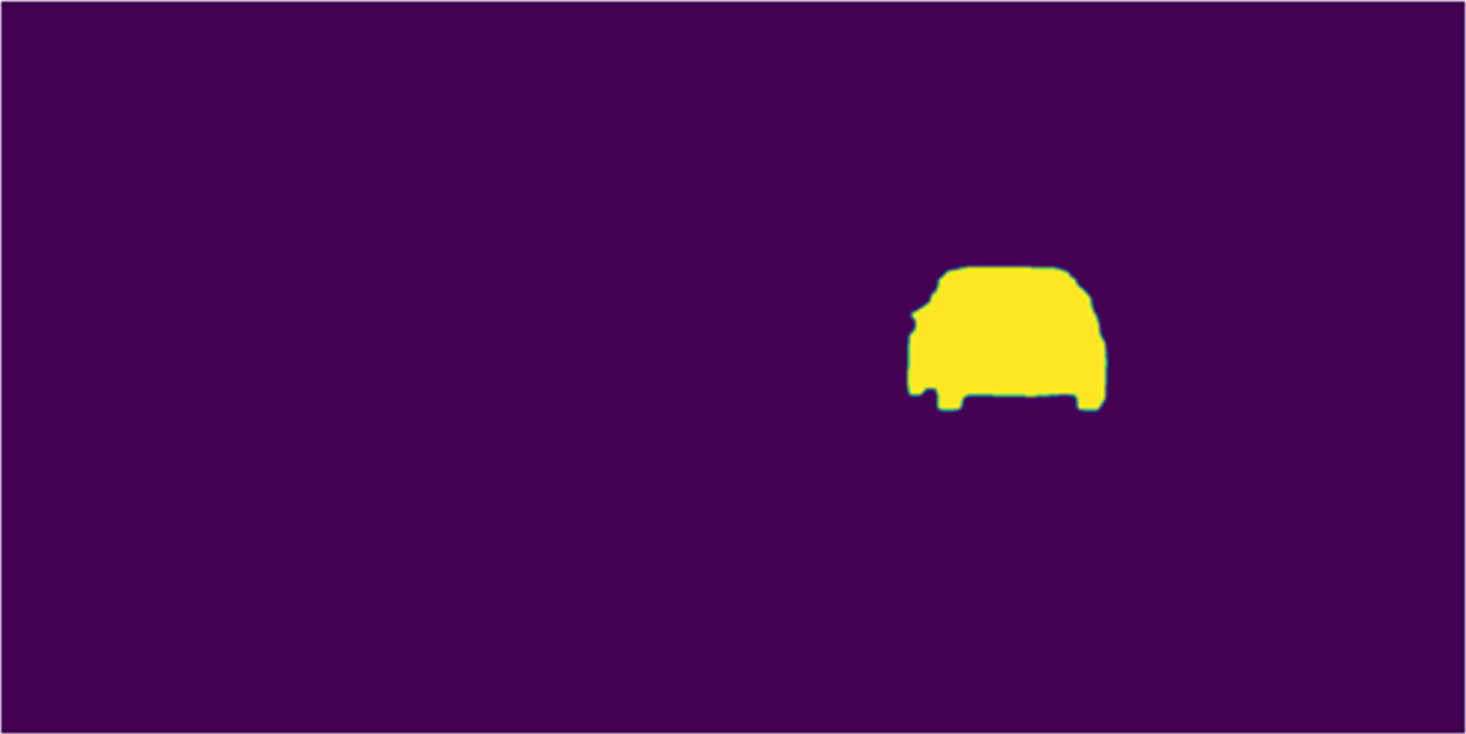}}
    \subfigure[$S_x$]{\includegraphics[width=0.3\linewidth]{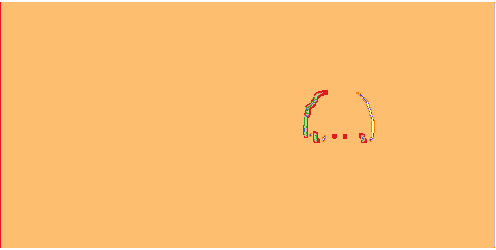}}
    \subfigure[$S_y$]{\includegraphics[width=0.3\linewidth]{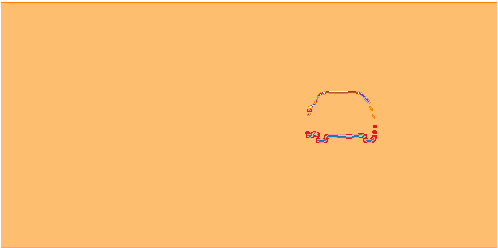}}
  \end{minipage}
  \caption{Edge information extracted by Sobel operator}
  \label{fig:three_images}
\end{figure}

\subsubsection{Laplacian Operator}
The Laplacian operator is a discrete filter in two dimensions, with the simplest isotropic derivative operator being the Laplacian operator. For a 2D digital image f(x,y), its Laplacian operator is defined by Equation (2): 
\begin{eqnarray}
\nabla^2f=\frac{\partial^2f}{\partial x^2}+\frac{\partial^2f}{\partial y^2}
\end{eqnarray}

The Laplacian operator can be implemented by convolving the convolutional kernel represented by $L$ in Equation (3). The operator can incorporate diagonal pixel information by integrating diagonal terms. In the following experiments, $L$ is chosen as the convolutional kernel for the Laplacian operator and we employ the it to extract the edge information of a specific instance.
\begin{eqnarray}
L=\begin{bmatrix}0&1&0\\1&-4&1\\0&1&0\end{bmatrix}
\end{eqnarray}
Laplacian's extraction results are showcased in Figure 3, revealing that the extracted edge lines of the instance's contour are slimer compared to the results obtained by the Sobel operator.

\begin{figure}[ht]
  \begin{minipage}[b]{\linewidth}
    \centering
    \subfigure[car]{\includegraphics[width=0.49\linewidth]{car.png}}
    \subfigure[edge extracted by Laplacian]{\includegraphics[width=0.49\linewidth]{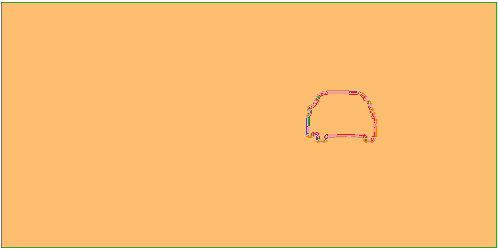}}
  \end{minipage}
  \caption{}
  \label{fig:two_images}
\end{figure}

\subsection{Fusion of the Edge Information}
We introduce an extra branch to the Yolact for predicting instance mask contours, this enhancement empowers the entire network to optimize instance edges. We name the improved model as RailYolact. The overall structure of the enhanced algorithm is depicted in Figure 5. Subsequently, we provide a comprehensive description of the specific implementation process of the RailYolact algorithm.
During the model's inference stage, the prototype masks are fused to produce predicted instance masks ($M_p$) for the objects in the input image. The instance masks for n objects in a single image can be represented as an n × w × h tensor. This n × w × h tensor serves as the input to the edge information prediction branch. In the previous section, we have thoroughly introduced two edge detection operators. By applying these operators to the predicted instance masks ($M_p$), we obtain the predicted instance edge masks ($M_{pedge}$). Similarly, by applying the edge detection operators to the ground truth instance masks ($M_{gt}$), we obtain the labels for the instance edge masks ($M_{gtedge}$). With the predicted and labeled values at hand, we can construct a loss function to optimize the disparity between $M_{pedge}$ and $M_{gtedge}$, enabling the network to optimize the instance edges during the inference process.
In this paper, we define the edge detection operator as K, which can be the Sobel operator or the Laplacian operator. The $M_{mask}$ represents instance mask. The extraction process of the instance edge mask can be formulated as Equation (4), where the operations is convolution operation.

\begin{eqnarray}
M_{edge}=K*M_{mask}
\end{eqnarray}

The structure of the edge information extraction head is shown in Figure 4. In this figure, n × W × H represents the masks of n instances, and 3 × 3 × D represents the edge detection operator. This operator can be implemented as a convolutional kernel in a convolutional neural network, with the parameters known and no bias terms.

\begin{figure}[ht]
  \centering
  \includegraphics[width=0.8\linewidth]{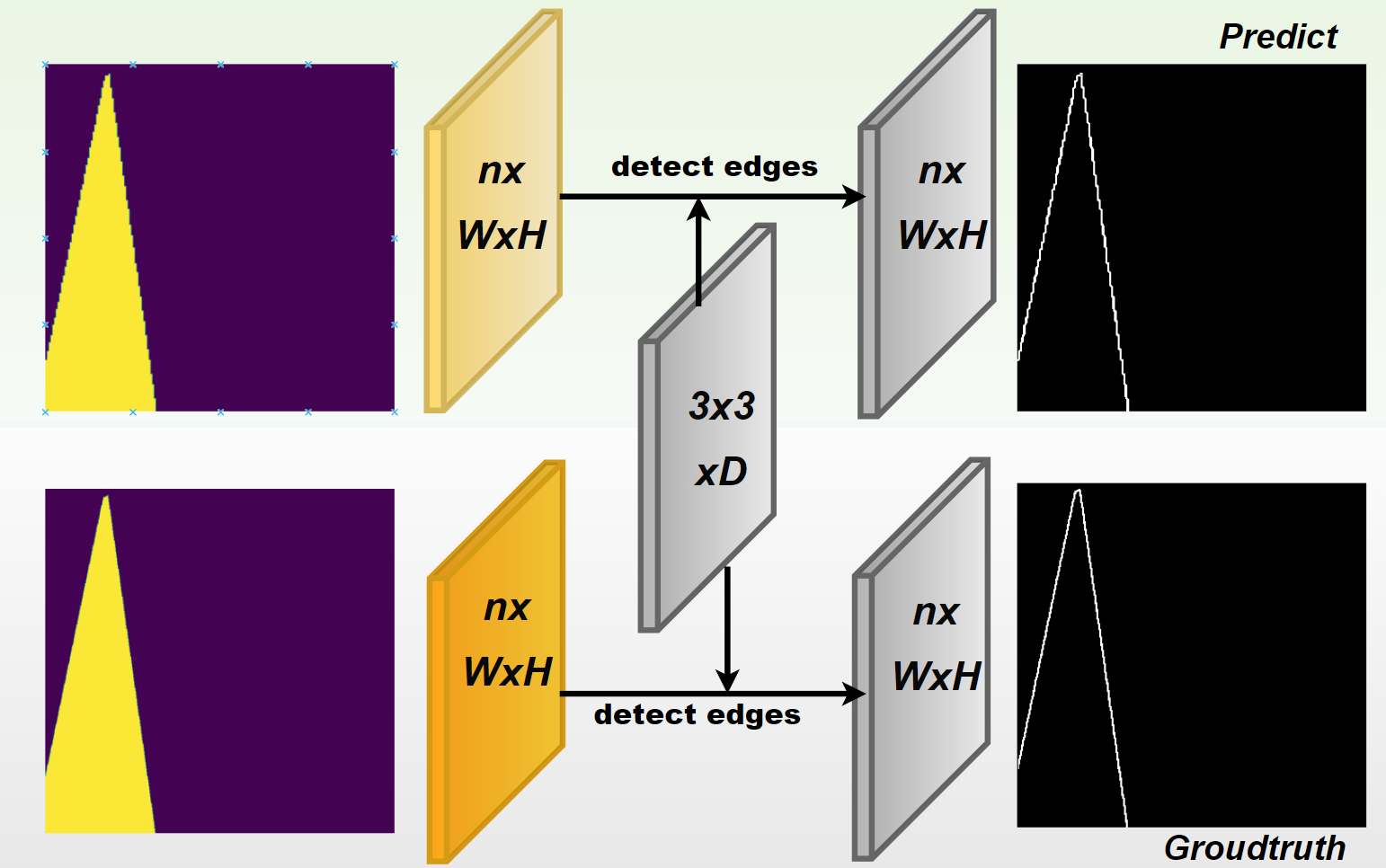}
  \caption{Edge information extraction head}
  \label{fig:images}
\end{figure}

\begin{figure*}[t]
  \centering
  \includegraphics[width=\textwidth]{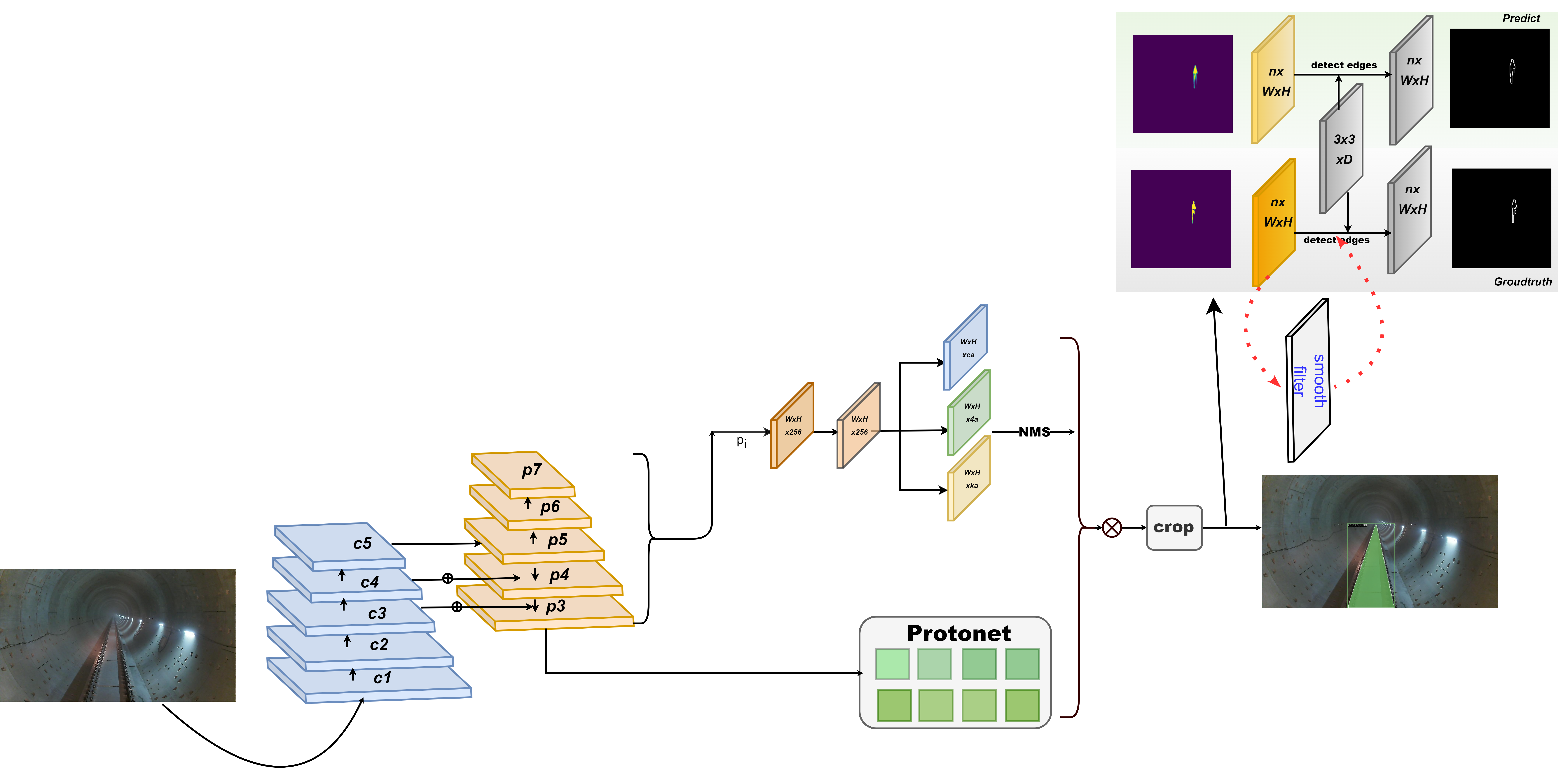}
  \caption{RailYolact}
  \label{fig:my-figure}
\end{figure*}
\subsection{Loss}
The model's loss is defined as a weighted sum of four types of losses: classification loss ($L_{cls}$), bounding box regression loss ($L_{bbox}$), instance mask loss ($L_{mask}$), and edge information loss ($L_{edge}$). The definitions of the first three losses are consistent with the Yolact paper. To compute the edge loss, we adopt the binary cross-entropy loss function, as shown in Equation (5), similar to the mask loss. This loss function allows us to measure the dissimilarity between the edge of the predicted masks and the true edge of the ground truth masks.
\begin{eqnarray}
L_{mask}=BCE(M_{gtedge},M_{pedge})
\end{eqnarray}
In the RailYolact model training process, the weights allocated to $L_{cls}$, $L_{bbox}$, $L_{mask}$ are 1, 1, 1.125, respectively. In the process of incorporating $L_{edge}$ into the Loss, the initial approach is directly adding to the Loss. However, this method did not yield satisfactory results. To establish a meaningful relationship between the mask loss and the edge map loss, $L_{edge}$ is ultimately defined as Equation (6).
\begin{eqnarray}
L_{edge}=L_{mask}*e^{\frac{L_{edge}}{4}}
\end{eqnarray}

\subsection{A Strategy to Eliminate Jagged Edges of Rail Masks}
During the training of the network, the size of the input image instance mask is set to 800 × 800, whereas the size of the masks generated by the entire algorithm for each instance is 200 × 200. In the process of calculating the mask loss, it is necessary to scale the actual labels to match the size of the predicted masks, typically achieved through linear interpolation. However, due to the substantial scaling factor employed, the labels themselves exhibit a jagged appearance, which the network inadvertently learns during the training process. Consequently, when the network performs inference, the segmented mask edges in the output display jaggedness. 

% We visualized the results of certain instance labels after undergoing interpolation, revealing the presence of jagged labels, as illustrated in Figure 6. The left of the figure showcases the real labels of the rail masks, while the right depicts the masks that have been scaled using linear interpolation.

% \begin{figure}[ht]
%   \begin{minipage}[b]{\linewidth}
%     \centering
%     \subfigure[ground truth mask]{\includegraphics[width=0.48\linewidth]{gt_masks0.png}}
%     \subfigure[interpolated mask]{\includegraphics[width=0.48\linewidth]{mask_targets0.png}}
%   \end{minipage}
%   \caption{}
%   \label{fig:two_images}
% \end{figure}

The box filter is the most basic type of low-pass filter, created by multiplying an array consisting of values all set to 1 by a normalized constant. It can be represented by Formula(7):

\begin{eqnarray}
K=\frac{1}{mm}\begin{bmatrix}1&\cdots&1\\\vdots&\ddots&\vdots\\1&\cdots&1\end{bmatrix}_{mm}
\end{eqnarray}

We employ box filtering to process the interpolated labels. As the instance masks are binary images, the pixel values remain unchanged within regions where the mask grayscale values are constant. Conversely, in areas with substantial pixel variations, the pixel values within a box filter kernel are smoothed, thereby mitigating the jagged instance edge. Box filtering is a simple implementation but can yields impressive smoothing outcomes. 

% Figure 7 provides a visual comparison before and after the smoothing process.

% \begin{figure}[ht]
%   \begin{minipage}[b]{\linewidth}
%     \centering
%     \subfigure[smooth targets]{\includegraphics[width=0.48\linewidth]{mask_targets1.png}}
%     \subfigure[smooth results]{\includegraphics[width=0.48\linewidth]{guidao2_heshi.png}}
%   \end{minipage}
%   \caption{}
%   \label{fig:two_images}
% \end{figure}

% The filter is applied at the position illustrated in Figure 8. 
Throughout network training, we utilize the smoothed masks as the ground truth labels for computing the loss with predicted masks. Since the smoothing filter only operates during the training process, it does not impact the inference speed. Although the smoothing edge processing leads to a marginal loss of information in the true mask labels, resulting in a slight influence on the segmentation accuracy, this trade-off is worth to eliminate the jagged effects at the edges.

% \begin{figure}[ht]
%     \centering
% \includegraphics[width=0.6\linewidth]{remove_jagged_edge.png}
%   \caption{Edge head structure}
%   \label{fig:images}
% \end{figure}

\section{Experiments}
We perform a thorough comparison of RailYolact to the state of the art along with comprehensive ablation experiments. We conducted experiments on our self-built railway dataset and Cityscapes~\cite{cordts2016cityscapes}, and evaluated the results using standard COCO metrics~\cite{lin2014microsoft} including AP (averaged over IoU thresholds) and $AP_{50}$.

\subsection{Railway Datasets}
The rail segmentation dataset is made by ourselves, it was collected by equipping video capture devices on both the front and rear of the rail vehicle. The captured videos were divided into individual frames at regular intervals. During the annotation process, the images were formatted to COCO dataset. After preprocessing, the dataset was divided into a training set of 7,275 images and a validation set of 3,120images, all with a size of 1280 × 720 pixels.

\subsection{Results on Railway Datasets}
In order to evaluate the pratical value of the method proposed in this paper in engineering applications, we applied the RailYolact to the rail dataset. This allowed us to achieve rapid and accurate segmentation of rail positions in images. Prior to being fed into the network, all images were resized to a uniform size of 800 × 800. The comparative results of Yolact and RailYolact on the rail segmentation dataset are summarized in Table 1, with S represents Sobel operator and L represents Laplacian operator. It is evident that using the Laplacian operator for extracting edge information yields superior results compared to using the Sobel operator in RailYolact, both in terms of FPS and AP.

\begin{table}[ht]
\begin{center}
\caption{comparison results of RailYolact and Yolact on railway dataset} \label{tab:cap}
\begin{tabular}{|c|c|c|c|}
  \hline
  Method & FPS & AP & $AP_{50}$
  \\
  \hline
  Yolact & 27.2 & 97.2 & 99.0\\
  RailYolact-S & 27.8 & 97.9 & 99.0\\
  RailYolact-L & 28.4 & 98.2 & 99.0\\
  \hline
  
\end{tabular}
\end{center}
\end{table}

Although the results presented in the table reveal minimal differences among Yolact and RailYolact in terms of AP evaluation metrics, there is a substantial improvement in the visual perception of the outcomes. Our analysis attributes this finding to the fact that each image in the dataset contains only a single rail, representing a large object instance of a single category. 

The original rail images is presented in Figure 6(a). The result of rail segmentation using Yolact is presented in Figure 6(b) , and the result of Yolact after adding edge information extracted by Laplacian operator is presented in Figure 6(c), the result of RailYolact using Laplacian operator is presented in Figure 6(d). It is apparent that the introduction of edge information improves the smoothness, accuracy, and trustworthiness of the segmented edges in the results. Applying the box filter effectively reduces the jagged edges in the predicted rail segmentation mask. 

\begin{figure}[ht]
  \begin{minipage}[b]{\linewidth}
    \centering
    \subfigure[Original]{\includegraphics[width=0.48\linewidth]{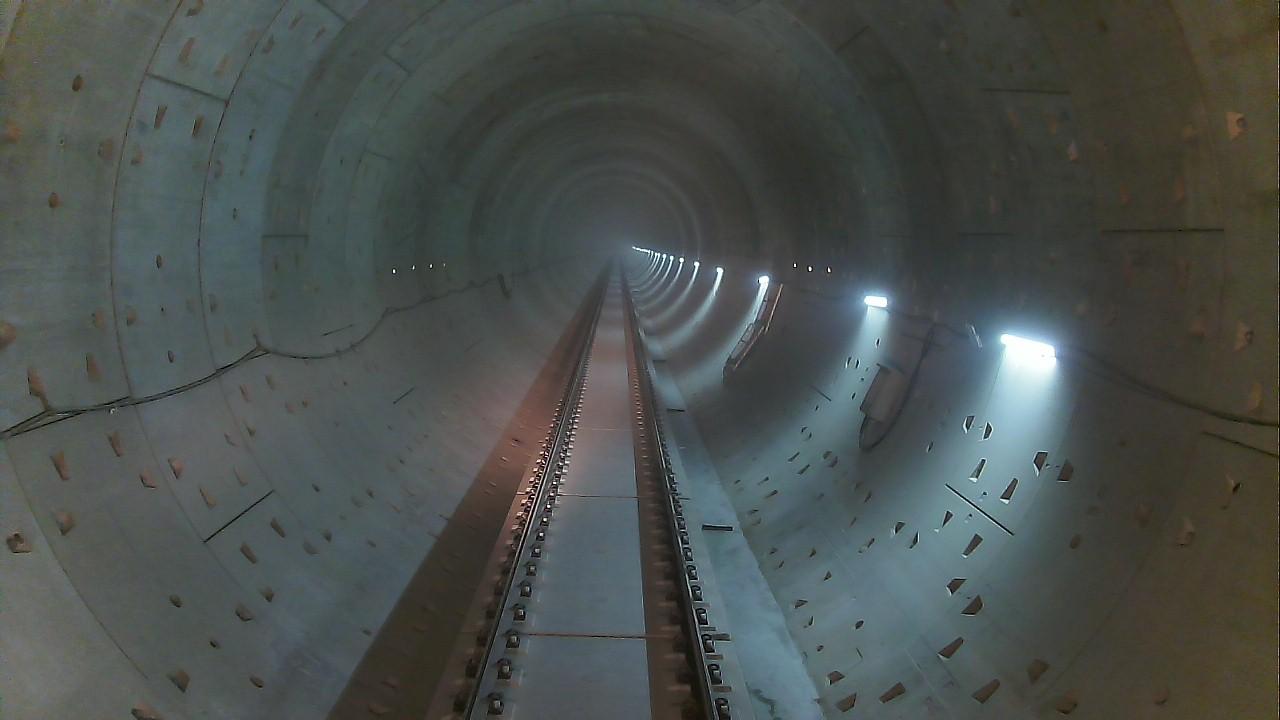}}
    \subfigure[Yolact]{\includegraphics[width=0.48\linewidth]{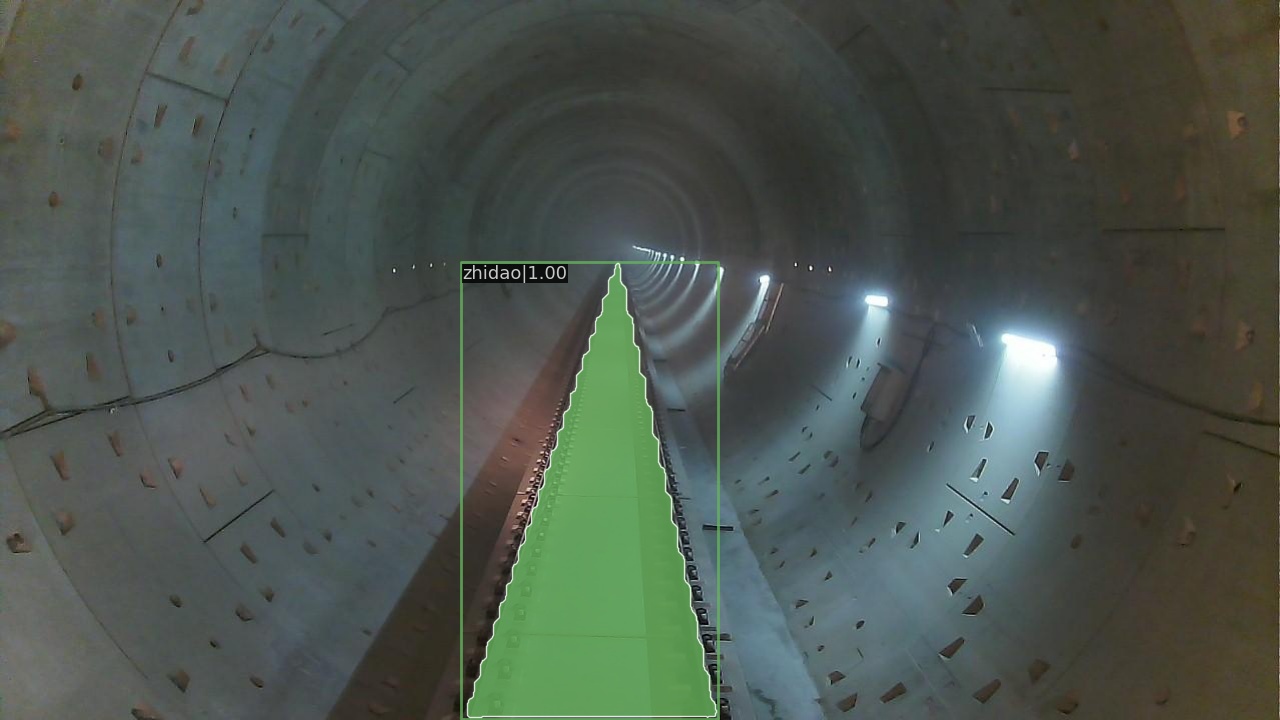}}
    \subfigure[Yolact with edge information]{\includegraphics[width=0.48\linewidth]{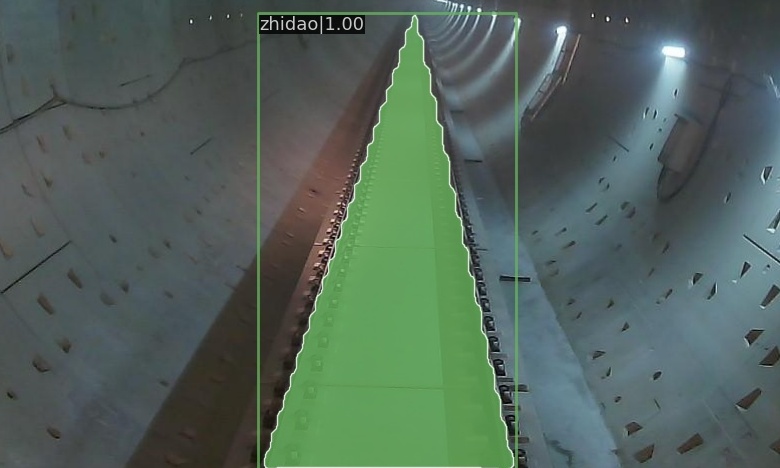}}
    \subfigure[RailYolact-Laplacian]{\includegraphics[width=0.48\linewidth]{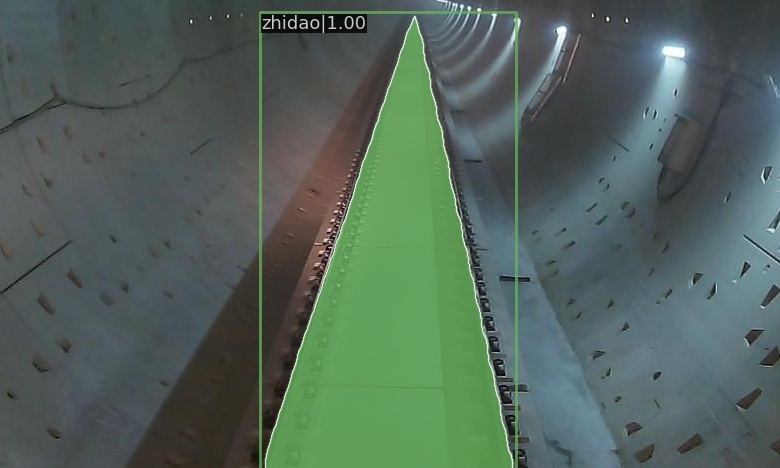}}
  \end{minipage}
  \caption{}
  \label{RailSegCompare}
\end{figure}

\subsection{Results on Cityscapes}
\subsubsection{Performance}
 In order to assess the impact of the improvement approach in RailYolact and its generalization ability, we conducted additional experiments on Cityscapes~\cite{cordts2016cityscapes}. Since instances in public datasets do not have edges as large as individual rail instances, the elimination of jagged edges in instance mask labels was not performed in this experiment. The comparative results of the two algorithms on Cityscapes~\cite{cordts2016cityscapes} are presented in Table 2, where RailYolact-L and RailYolact-S representing RailYolact using the Laplacian and Sobel operators, respectively.The R means Resnet.

\begin{table}[ht]
\begin{center}
\caption{} \label{tab:cap}
\begin{tabular}{|c|c|c|c|c|}
  \hline
  Method & Backbone & FPS & AP & $AP_{50}$
  \\
  \hline
  Yolact & R-50-FPN & 26.8 & 19.8 & 36.2\\
  RailYolact-L & R-50-FPN & 28.4 & 23.9 & 40.8\\
  RailYolact-S & R-50-FPN & 27.4 & 23.6 & 41.9\\
  RailYolact-S & R-101-FPN & 21.1 & 23.1 & 41.2\\
  \hline
  
\end{tabular}
\end{center}
\end{table}

From Table 2, it is clear that the our method achieves a 4.1 point increase in AP compared to the Yolact. Importantly, this performance enhancement does not entail any additional speed loss since the integration of edge information does not occur during network inference. An intriguing observation from Table 2 is that the RailYolact experiences a slight increase in inference speed when it exhibits better performance. This phenomenon can be attributed to the optimization effect of the entire model, where certain irrelevant weight values in the network may be eliminated, such as some convolutional kernels be sparsed, leading to a slight enhancement in the network's inference speed.

Additionally, it can be observed that different edge detection operators have similar effects on the model. Furthermore, to compare the differences between different backbone feature extraction networks, we performed experiments comparing ResNet50 and ResNet101. On ResNet101, the overall AP of the model showed relatively poor performance. This could be attributed to the relatively small scale of Cityscapes~\cite{cordts2016cityscapes} and the limited number of instances per category in the dataset.

% The results from Table 3 indicate that when the backbone feature extraction network is ResNet101, the overall model performs better on small instances but relatively worse on larger instances like train. We speculate that this may be related to the selection of input features for the network.

% \begin{table*}[ht]
% \begin{center}
% \caption{} \label{tab:cap}
% \begin{tabular}{|c|c|c|c|c|c|c|c|c|c|}
%   \hline
%   % after \\: \hline or \cline{col1-col2} \cline{col3-col4} ...
%   Method & Backbone & Person & Rider & Car & Truck & Bus & Train & Motorcycle & Bicycle
%   \\
%   \hline
%   Yolact & R-50-FPN & 7.7 & 6.0 & 34.0 & 27.0 & 44.7 & 21.0 & 6.5 & 7.6\\
%   RailYolact-L & R-50-FPN & 8.6 & 6.6 & 40.8 & 31.7 & 52.0 & 42.1 & 8.2 & 7.9\\
%   RailYolact-S & R-50-FPN & 8.6 & 7.1 & 41.9 & 32.6 & 50.3 & 41.8 & 8.2 & 8.1\\
%   RailYolact-S & R-101-FPN & 9.2 & 6.4 & 41.2 & 32.7 & 51.4 & 33.5 & 9.1 & 8.5\\
%   \hline
  
% \end{tabular}
% \end{center}
% \end{table*}

To compare the performance of RailYolact with other algorithms in terms of segmentation mask accuracy and inference speed, we conducted a comparative analysis of the AP and inference speed of our algorithm and other algorithms on Cityscapes~\cite{cordts2016cityscapes} using one RTX3080. Since most algorithms have only been tested on the MScoco dataset, we only compared our algorithm with the algorithms mentioned in the Mask R-CNN~\cite{he2017mask} paper, as shown in Table 3, due to some algorithms focusing only on improving accuracy and disregarding speed, we did not gather statistics on their FPS. It can be observed that the network models trained with our algorithm have a significant advantage in terms of speed compared to other two-stage algorithms. This makes them highly valuable for engineering applications that require fast inference speed in simple scene scenarios.

\begin{table}[ht]
\begin{center}
\caption{} \label{tab:cap}
\begin{tabular}{|c|c|c|c|}
  \hline
  Method & FPS & AP & $AP_{50}$
  \\
  \hline
  Yolact & 27.2 & 19.8 & 36.2\\
  RailYolact & 28.4 & 23.8 & 40.9\\
  InstanceCut~\cite{kirillov2017instancecut} &  & 13.0 & 27.9\\
  DWT~\cite{bai2017deep} &  & 15.6 & 30.0\\
  SAIS~\cite{de2017semantic} &  & 17.4 & 36.7\\
  DIN~\cite{arnab2017pixelwise} &  & 20.0 & 38.8\\
  SGN~\cite{liu2017sgn} &  & 25.0 & 44.9\\
  Mask R-CNN~\cite{he2017mask} & 8.6 & 32.0 & 58.1\\
  % InstanceCut &  & 13.0 & 27.9\\
  % DWT &  & 15.6 & 30.0\\
  % SAIS &  & 17.4 & 36.7\\
  % DIN &  & 20.0 & 38.8\\
  % SGN &  & 25.0 & 44.9\\
  % Mask R-CNN & 8.6 & 32.0 & 58.1\\
  \hline
  
\end{tabular}
\end{center}
\end{table}

\section{CONCLUSION}
In this paper, we proposed the RailYolact emphasizing instance edge information and implement it for rail segmentation. We employ edge operators to calculate the edge information of the predicted rail masks and the annotated ground truth rail masks. The calculated edge information loss is then incorporated into the loss function to enhance the model's emphasis on edge of the rail. Besides, to solve the problem of jagged edges in the rail segmentation results, we use box filters to resolve the jagged edges of the mask caused by linear interpolation of ground truth mask. Ablation experiments conducted on our custom rail dataset and Cityscapes~\cite{cordts2016cityscapes} demonstrate that this improvement enhances prediction accuracy while preserving the real-time nature of the model's inference. The model we proposed can also be applied to other tasks that need to meet real-time requirements and are sensitive to edge information.

\bibliographystyle{IEEEbib}
\bibliography{icme2023template}

\end{document}